# Predicting Elevated Risk of Hospitalization Following Emergency Department Discharges

Dat Hong[1], Philip M. Polgreen[2], Alberto Maria Segre[1]

[1]Department of Computer Science, [2]Department of Internal Medicine
The University of Iowa
Iowa City, Iowa (USA)
{dat-hong, philip-polgreen, alberto-segre}@uiowa.edu

*Abstract*—Hospitalizations that follow closely on the heels of one or more emergency department visits are often symptoms of missed opportunities to form a proper diagnosis. These diagnostic errors imply a failure to recognize the need for hospitalization and deliver appropriate care, and thus also bear important connotations for patient safety. In this paper, we show how data mining techniques can be applied to a large existing hospitalization data set to learn useful models that predict these upcoming hospitalizations with high accuracy. Specifically, we use an ensemble of logistics regression, naïve Bayes and association rule classifiers to successfully predict hospitalization within 3, 7 and 14 days of an emergency department discharge. Aside from high accuracy, one of the advantages of the techniques proposed here is that the resulting classifier is easily inspected and interpreted by humans so that the learned rules can be readily operationalized. These rules can then be easily distributed and applied directly by physicians in emergency department settings to predict the risk of early admission prior to discharging their emergency department patients.

## I. INTRODUCTION

Medical errors are an important source of morbidity and mortality [1, 2]. Thus reducing errors is an important patient safety goal and public health priority [1-3]. An increasingly recognized source of medical errors are delayed diagnoses and other diagnostic-related errors [4-7]. Delayed diagnoses and diagnostic errors can lead to lead to not only excess healthcare costs, but also negative and even catastrophic outcomes for patients [8-10]. Diagnostic delays and other diagnostic-related errors may be especially common in emergency departments given the high patient loads, fast pace of patient turnover with limited time for observation, the general lack of continuity of care, a lack of detailed information regarding the patient, and the large range of patient acuity. Indeed, analysis of malpractice claims demonstrate the frequency and importance of diagnostic delays and errors in emergency departments [11, 12].

One potential marker for quality of care in emergency departments is unanticipated short-term revisits [13]. Approximately 7% of patients seen in emergency departments return in 3 days and 22% within 30 days [14]. While some of those returns may be expected or unavoidable, some of the returns are associated with diagnostic and other errors such as insufficient treatment or care [13]. Thus, excess revisits have long been proposed as a marker of quality for emergency department care [15-17]. However, not all revisits are unanticipated, nor do the vast majority represent the provision of inadequate medical care [18, 19]. Some revisits may be scheduled, and some may be due to patient-related factors, such as a lack of understanding, anxiety, or progression of disease [18, 20]. An estimated 5-20% of revisits are attributable to issues related to the possibly less-than-optimal care provided at the index emergency department visit [19]. Not surprisingly, hospitalizations following discharge from the emergency room are a better marker for tracking quality of care than simply noting return ED visits without hospitalization [21, 22].

While several reports have focused on *risk factors* associated with returns to the emergency department [23-35], fewer have focused on *predicting* revisits to the emergency department or predicting hospitalizations following emergency discharges [36-39]. Among the work that has been done, much has been focused older populations, pediatric populations or specific conditions, limiting their generalizability.

Given that hospital admissions following patients' discharges from the emergency department are undesirable, the goal of this paper was to predict patients at high risk for being admitted to the hospital at either 3, 7 or 14 days following a discharge from an emergency department using a large population-based sample over a large geographic region.

## II. METHODS

### A. Diagnostic Errors

There are many definitions of missed diagnoses and/or missed opportunities. In this paper we consider diagnoses that are "missed, wrong, or delayed, as detected by some subsequent definitive test or finding" [40]. We use a hospital admission within a fixed time window (3, 7 or 14 days) after an emergency department (ED) visit as the "subsequent definitive test" and select patients of interest by looking back in time from these "index" admissions. The 3, 7 and 14 day windows are suggested by related work on hospital readmissions [41].

### B. Dataset

The data used here were extracted from the 2009 Healthcare Cost and Utilization Project (HCUP) California inpatient

Support for this research was provided in part by University of Iowa Health Ventures' Signal Center for Health Innovation.

database (SID) and emergency department database (SEDD). The former contains records of all inpatient discharges from short-term acute care non-federal-government hospitals in California, while the latter contains records of emergency department visits that do not lead to immediate hospitalizations. Each record includes the principal and secondary diagnoses, procedures performed, demographic information, length of stay, admission and discharge status, hospital charges and payment sources. California HCUP data patient SID and SEDD records can be linked across visits using an anonymized patient record index (*visitlink*) and simple calendrical calculations relating the implied visit dates (*daystoevent*), producing a comprehensive view of patient ED visits and hospitalizations across facilities and over time.

The 2009 California SEDD dataset contains 9,875,973 deidentified ED visits. We remove records that lack patient identifiers (these cannot be linked to the SID data) as well as records pertaining to pregnant women (a primary CCS code between 177 (spontaneous abortion) and 196 (other pregnancy and delivery including normal) for at least one ED or hospital visit) and children (age < 18), producing a final dataset for analysis consisting of 5,487,722 ED visits. Each ED visit is considered a diagnostic error if a hospital admission occurs within 7, 14 or 30 days of an ED visit, excluding hospital admissions for mental illness (an outsize number of readmissions are due to CCS codes associated with mental illness, between 650 and 670). The resulting dataset (see Figure 1) is markedly imbalanced, with only 2% of the records in the dataset having a qualifying hospital readmission within a 14-day window (and concomitantly smaller percentages for 3 and 7-day windows).

*C. Feature Selection*

Each dataset record consists of 152 features, including age, admission date, race, length of stay, patient disposition code, and so on. 102 of the features correspond to diagnoses, procedure, and injury codes in various coding formats (ICD-9 and CCS for diagnoses, CPT-4/HCPCS and CCS for procedures, and ICD-9-CM and CCS for injury codes). For this study, we retain the 21 CCS (Clinical Classification Software) coded diagnosis fields and the 21 CCS-coded procedure fields, dropping the (redundant) CPT-4/HCPCS/ICD-9 coded fields [42, 43]. The remaining 106 features (56 CCS-coded fields and 50 other features) are then dummy coded (*e.g.,* the original admission month feature's twelve possible values would be recoded as 12 individual binary variables). Recoding features as binary dummy variables is a regression friendly strategy akin to binning when the size of the dataset is very large.

Each record is then augmented with some additional temporal information gleaned from the context of each visit. More specifically, each record receives the following additional features:

- Number of ED visits within last 30 days
- Number of ED visits to same facility within last 30 days
- Number of hospital visits within last 30 days
- Number of hospital visits to same facility within last 30 days
- Number of additional ED or hospital visits with same primary CCS code within last 30 days (default 0)
- The most frequent primary CCS code within the last 30 days.

Each record in the dataset thus contains a total of 3831 binary features.

*D. Classification Algorithms*

We wish to learn to recognize diagnostic errors as defined above based on the features just described. Our approach uses a weighted combination, or an *ensemble*, of instances of three types of learning algorithms: *logistic regression, naïve Bayes* and *association rule classification*.

Logistic regression (LR) is a commonly used technique that learns an estimator for the probability of a binary response from data. A LR classifier is a weighted linear combination of terms, where each term corresponds to an input feature and the weights for each term are fit from training data. Here, LR is provided access to all 3831 dummy coded features for each record in the appropriately sized window (3, 7 or 14 day) prior to a candidate readmission and fit to predict the probability of readmission. Among the primary advantages of LR are its simplicity and direct interpretability: simple inspection reveals which features are most important, as their weights will be larger than those of other features. LR has been used extensively in prior work to predict hospital readmission [44-46].

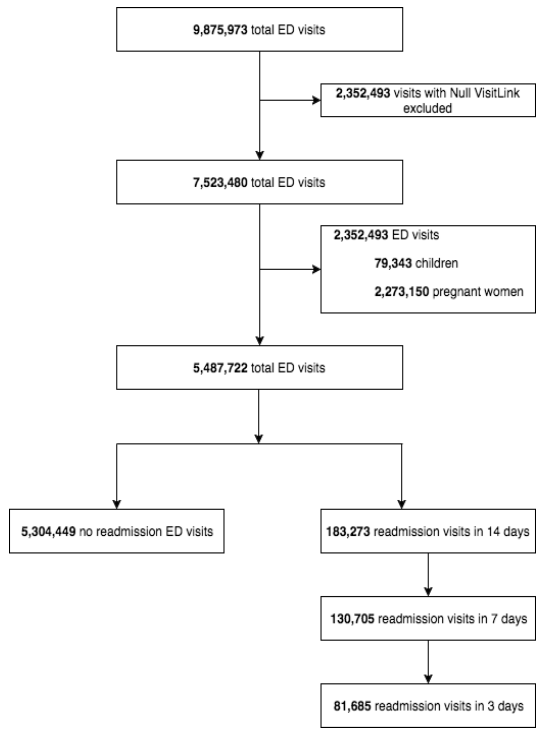

**Figure 1** Overview of the dataset: A readmission visit is a non-mental-health hospitalization occurring following at least one ED visit within the specified time window (excluding children under 18 and pregnant women).

Naïve Bayes (NB) is another commonly used supervised learning algorithm that assumes each input feature is independent from the other inputs (the "naïve independence" assumption). NB can be trained efficiently and is known to be a surprisingly good classifier [47]. Here, NB is provided the same inputs as LR and also asked to predict the probability of readmission.

Association rule classification (ARC) is a heuristic algorithm that uses association rule mining to identify sufficiently ``interesting'' combinations of features (or *itemsets*) based on frequency of co-occurrence (or *support*). Once appropriate itemsets have been identified, a heuristic is used to generate classification rules based on a selection of itemsets. In this study, we use the Apriori algorithm [48] to generate all frequent itemsets for rule generation; an example should help make this clear.

Consider the following simple example consisting of 7 records, each characterized by an itemset and a Boolean outcome (*i.e.,* 0 or 1; see Table 1). Recall an itemset is a set of items T ⊆ I, where I is the set of all features appearing in the dataset (I = <A,B,C,D,E> in this example). A rule has the form T => x, where x is the outcome. For each such rule, we can compute a corresponding *confidence*, which is the probability a record whose itemset subsumes the rule's antecedent has a matching outcome, and *support,* which is the number of records in the dataset whose itemset subsumes the rule's antecedent itemset. In this example, the rule antecedent itemset <A, B> has support 3 (*i.e.,* matches 3 records in the dataset); the rule <A, B> => 1 has confidence 0.66 (*i.e.,* the outcome 1 occurs in 2 of the 3 matching records) and the rule <A, B> => 0 has confidence 0.33 (*i.e.,* the outcome 0 occurs in 1 of the 3 matching records).

The algorithm proceeds in two steps: first, we construct all rules that satisfy a specified minimum support threshold, and, second, we build a classifier that applies these rules in decreasing order of support.

Rules are constructed by considering all elements in the powerset of I as possible rule antecedents. We construct a new rule for each candidate antecedent that appears in the training data by pairing it with each corresponding outcome, keeping the resulting rule only if it meets the prespecified support criteria. For the example shown in Table 1, a minimum support threshold of 2 produces the rules shown in Table 2.

TABLE 1

| ID | F1 | F2 | F3 | Class |
|----|----|----|----|-------|
| 1 | A | | | 0 |
| 2 | B | | C | 0 |
| 3 | A | B | D | 1 |
| 4 | B | E | | 1 |
| 5 | B | | A | 1 |
| 6 | C | B | | 0 |
| 7 | B | A | C | 0 |

TABLE 2

| Antecedent itemset | Support | Confidence (Output=0) | Confidence (Output=1) |
|---|---|---|---|
| <A> | 4 | 50% | 50% |
| <B> | 6 | 50% | 50% |
| <C> | 3 | 100% | 0% |
| <A,B> | 3 | 33% | 66% |
| <B,C> | 3 | 100% | 0% |

When presented with a new record for classification, we use the outcome associated with the most specific matching rule (*i.e.,* the rule having the largest matching antecedent, where matching means subsumed by the record in question), preferring the rule with higher confidence in the event of a tie. For this example, a new record <A, B, C, D> would match both rule antecedents <A, B> and <B, C>, with the classifier preferring the former thanks to its higher support. The classifier would then predict an outcome of 1 with probability 0.67, and an output of 0 with probability 0.33.

Because candidate rule generation can be expensive, the two association rule classifiers use a markedly reduced set of features drawn from the CCS codes alone (compare with the full dummy coded feature set provided to LR and NB). Moreover, ARC1 and ARC2 differ from one another in the itemsets they use to generate candidate rules. Here, ARC1 uses a "longitudinal" (or "vertical") itemset consisting only of primary CCS codes culled from ED or hospital visits in the 30 day period prior to admission. In contrast, ARC2 uses an "instantaneous" (or "horizontal") itemset consisting of all primary and secondary CCS diagnostic codes associated with the current visit alone. Note also that, for both ARC classifiers, we must choose an appropriate support threshold. If the threshold is set too low, there might be too many rule candidates; if it is set too high, we might miss a number of potentially valuable rules. Here we set the support threshold empirically so as to maximize performance on the training data.

*E. Ensemble Learning and Classification*

For the work reported here, we combine the outputs of four separate classifiers (LR, NB, ARC1 and ARC2), where each classifier is learned or fit independently. The outputs of these classifiers are then used as the inputs to an additional LR classifier that computes a linear combination of these outputs to produce the final classification. The coefficients in this second stage LR are also fitted from the data.

We proceed as follows. First, the original data set is randomly partitioned into three disjoint subsets. The first subset, consisting of 80% of the sample, are the *training data*, which are used to train each of the four classifiers. These same training data are also used to empirically set the support thresholds for ARC1 and ARC2 (10 and 40, respectively). The second subset, consisting of 10% of the sample, are the *validation data,* which are used to train the ensemble classifier. The third subset, consisting of the remaining 10% of the sample, are the *test data,* which are used to evaluate the overall performance of the ensemble.

*F. Measuring Performance*

Because the data available are extremely unbalanced (only 130,705 of 9,875,973 visits, or about 1.3%, of the 7-day time window data correspond to readmissions), it is not possible to use accuracy to measure performance of any of the classifiers. Instead, we report the area under the ROC (receiver operating characteristic) curve (AUC) as a measure of discrimination, defined as the ability to correctly classify missed diagnostic opportunities.

TABLE 3

| Window | LR | ARC1 | ARC2 | NB | Ensemble |
|---|---|---|---|---|---|
| 3-day | 0.7990 | 0.7218 | 0.7201 | 0.7564 | 0.8019 |
| 7-day | 0.8032 | 0.7244 | 0.7186 | 0.7575 | 0.8119 |
| 14-day | 0.8102 | 0.7320 | 0.7412 | 0.7577 | 0.8155 |

III. RESULTS

Table 3 reports the results obtained by each individual classifier as well as the ensemble overall for each of the 3-day, 7-day and 14-day window cases. In the results reported here, longer windows generally correspond to better AUC values, although this need not necessarily always be the case. Note also that the ensemble method always outperforms the individual classifiers (see Figure 2).

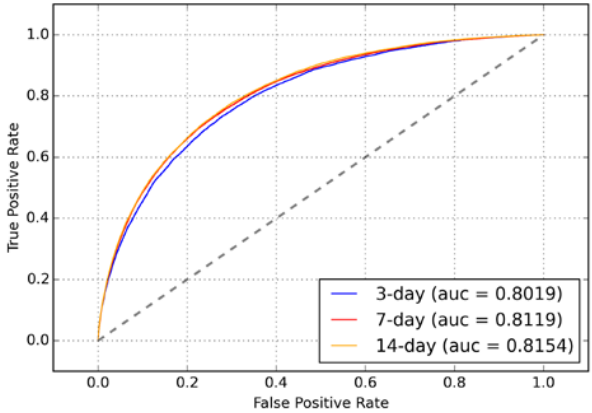

**Figure 2** AUC results for ensemble classifier using different time windows.

Examining the coefficients in the individual LR model yields some insight into which features can best be used to predict readmission. As might be expected, there is significant overlap in these important features across different time windows (3, 7 or 14 days). For example, the number of previous hospital visits tends to be an important predictor of readmission, where patients with frequent hospital visits prior to the ED visit will have a higher probability of readmission. In a similar fashion, certain CCS codes (*e.g.,* certain types of cancers, sickle cell anemia, encephalitis, cystic fibrosis, etc.) when they appear as the primary CCS code associated with a given ED visit are also associated with higher probability of readmission, as are certain injuries (*e.g.,* certain falls, aspirated foreign objects, open wounds, etc.). Coefficient inspection can also lend insight into coding idiosyncrasies: the procedure code for fetal monitoring was found to be associated with near-certain readmission in data that had supposedly excluded pregnant women. Closer examination revealed that 28% of ED visits coded for fetal monitoring were not also coded for pregnancy, and hence had not been removed.

TABLE 4

| ARC1 CCS | Support | Confidence |
|---|---|---|
| 50, 141, 250, 251 | 86 | 0.72 |
| 50, 141, 250 | 142 | 0.67 |
| 99, 102, 133 | 62 | 0.59 |
| 50, 141, 251 | 253 | 0.58 |
| 127, 157 | 52 | 0.58 |

| ARC2 CCS | Support | Confidence |
|---|---|---|
| 2, 237 | 63 | 0.54 |
| 211, 248 | 95 | 0.42 |
| 49, 248 | 102 | 0.42 |
| 122, 244, 246 | 65 | 0.41 |
| 114, 248 | 79 | 0.41 |

The performance of the ARC1 and ARC2 classifiers is also quite interesting. ARC1 obtains AUC values that are comparable to NB by matching its rule antecedents against only the primary CCS codes culled from records in the appropriately sized window prior to the ED visit in question. ARC2 obtains results of nearly similar quality in terms of AUC by matching its rule antecedents against the primary and secondary CCS codes for the current ED visit alone. Furthermore, like for LR, ARC1 and ARC2 deal with directly interpretable features. By looking at the rules with high confidence and support, we know, for example, what combination of CCS codes in a given ED visit (ARC2) are associated with increased probability of readmission. For our data, a CCS code of 248 (gangrene) when co-occurring with 49 (diabetes mellitus without complication), 114 (peripheral and visceral atherosclerosis), or 211 (other connective tissue disease) is a good indicator of readmission. For ARC1's longitudinal view of primary CCS codes, visits with primary CCS codes of 50 (diabetes mellitus with complications) and 141 (other disorders of the stomach or duodenum) with either 250 (nausea and vomiting) or 251 (abdominal pain) are similarly associated with high rates of readmission (Table 4 shows top 5 association rule antecedents for both ARC1 and ARC2).

IV. DISCUSSION

Our results show that we can, using only administrative data which is readily and widely available, predict, with a reasonably degree of certainty, which patients are likely to be hospitalized after leaving the emergency department at 3, 7 and 14 days. Our results compare favorably to other prediction attempts of revisits, hospitalizations or re-hospitalizations may of which have longer time horizons (*e.g.,* 30 days) [25, 38, 49-55]. Perhaps not surprisingly, some efforts to predict hospital admissions at the time of the index emergency department visit at the time of triage (first presentation) using administrative data have been more successful, ROC = 0.85 [56]. Our approach, unlike some other machine learning approaches, is relatively interpretable and can be used to selectively help physicians and other healthcare professionals make decisions about whether to (1) admit patients instead of sending them home from the emergency department or (2) allocate resources after a patient leaves the emergency department to help prevent future hospitalizations.

Understanding the transition between emergency department visits and hospitalizations is extremely important given that over 80% of all unscheduled hospital admissions in the United States originate from emergency department visits [57]. In fact, the proportion of admissions originating from the ED has increased dramatically over the last several years [57]. The increase in emergency room use has led to overcrowding of emergency departments which in turn affects quality of care and patient outcomes [58]. To prevent revisits and bounce backs from emergency departments, multiple approaches have been applied, including telephone calls following patients discharged to home by nurse navigators and attempts to schedule primary care visits for patients discharged to home. What is needed is an approach to allocate such post-discharge interventions effectively. Many patients in the emergency department are not likely to need such interventions. We believe our approach, by targeting patients that are likely not only to return, but get admitted to the hospital, provides a promising target for quality and patient satisfaction initiatives. And while many projects investigate revisits to the emergency department, these projects are descriptive, and focus on identifying risk factors rather than making predictions.

Our work can also be extended into a decision-support system. Given that the data we use is generally available, our outcome results, based on association rules, are easy to interpret. While the claims data are not available at the time of decision to discharge, all of the information, in theory, that is used to generate the discharge data is available. In future work, we need to validate our results in other large geographic regions (*e.g.,* states other than California) and with other years. Such work is underway. Ultimately, our decision-support work could help physicians make more informed choices about patients on the margin (i.e., patients who for whom it is not immediately clear if they should be admitted to the hospital or sent home. Patients who are discharged to home but are flagged as high risk for a future admission could be targeted for close follow up with a visit to their primary care, a call from a nurse or pharmacist. Indeed, not all patients, even if they are told they are likely to be admitted in the future, may wish to be admitted, and some may prefer a trial of care at home, prior to an admission [41] .

Our paper has several limitations. First, we use administrative data exclusively and do not include specific observable data that may be important to determine patient severity when presetting to an emergency department (e.g. vital signs, medications, triage assessment from notes). Second, our data are only for the state of California. It is possible, but unlikely, that patients seen in California emergency departments are admitted to a hospital in another state, and we would not know about this. Third, some information that may be important for predicting visits to the emergency department are not in our data, for example, health literacy or language difficulties (*e.g.,* patients for whom English is not their primary language) [59]. In addition, we have limited information about the hospitals and emergency departments, and institutional values like teaching status and size may affect patient outcomes. Some of this information could be added to our model by linking our data to information from, for example, the American Hospital Association database. Fourth, we did not focus on admissions to the hospital in more than 14 days. Historically, some quality metrics look at revisits to emergency departments within 72 hours, although there is no empirical evidence for looking at this period [14]. A 7-day period may be a more reasonable period for measuring quality of care in emergency departments [60], because revisits to an emergency department within 7 days are most likely to be related to the same health problem as the index visit [30]. For this project we chose 7 and 14 days based on a work that identified revisits at 9 days following an initial emergency department based on a "time-to-return-curve" analysis for identifying potentially avoidable re-visits to the emergency department [14]. Finally, one can always learn from larger datasets, or use more features (recall only CCS codes were used in ARC1 and ARC2) when training the underlying learning algorithms.

Despite these limitations, the results from this pilot study demonstrate that using only a large administrative database, we can develop models that can help predict which patients, after leaving the emergency department, are most likely to be admitted to a hospital either within a 3-, 7- or 14-day period. This approach can be used to allocate scarce resources such as calls from nurse navigators and pharmacists. However, it can also be used to investigate new quality metrics and ultimately inform the building of diagnostic support tools to automatically flag high-risk patients. Finally, because our approach, unlike some machine-learning approaches, which operate like a "black box", leads to associate rules that are easy to interpret, we may learn of novel risk factors and combinations of factors, accounting for the ordering of events that would be much more difficult to discover using traditional epidemiological methods.

V. REFERENCES

1. Phillips, D.P. and C.C. Bredder, *Morbidity and mortality from medical errors: an increasingly serious public health problem.* Annual Review of Public Health, 2002. **23**(1): p. 135-150.
2. Clancy, C., *Improving patient safety-five years after the IOM report.* The New England journal of medicine, 2004. **351**(20): p. 2041.
3. Leape, L.L., et al., *Promoting patient safety by preventing medical error.* Jama, 1998. **280**(16): p. 1444-1447.
4. Zwaan, L. and H. Singh, *The challenges in defining and measuring diagnostic error.* Diagnosis (Berl), 2015. **2**(2): p. 97-103.
5. Newman-Toker, D.E. and P.J. Pronovost, *Diagnostic errors--the next frontier for patient safety.* Jama, 2009. **301**(10): p. 1060-2.
6. Wachter, R.M., *Why diagnostic errors don't get any respect--and what can be done about them.* Health Aff (Millwood), 2010. **29**(9): p. 1605-10.
7. Singh, H., *Diagnostic errors: moving beyond 'no respect' and getting ready for prime time.* BMJ Qual Saf, 2013. **22**(10): p. 789-92.
8. Graber, M.L., N. Franklin, and R. Gordon, *Diagnostic error in internal medicine.* Arch Intern Med, 2005. **165**(13): p. 1493-9.
9. Schiff, G.D., et al., *Diagnostic error in medicine: analysis of 583 physician-reported errors.* Arch Intern Med, 2009. **169**(20): p. 1881-7.
10. Saber Tehrani, A.S., et al., *25-Year summary of US malpractice claims for diagnostic errors 1986-2010: an analysis from the National Practitioner Data Bank.* BMJ Qual Saf, 2013. **22**(8): p. 672-80.
11. Kachalia, A., et al., *Missed and delayed diagnoses in the emergency department: a study of closed malpractice claims from 4 liability insurers.* Ann Emerg Med, 2007. **49**(2): p. 196-205.
12. White, A.A., et al., *Cause-and-effect analysis of risk management files to assess patient care in the emergency department.* Acad Emerg Med, 2004. **11**(10): p. 1035-41.
13. Verelst, S., et al., *Short-term unscheduled return visits of adult patients to the emergency department.* J Emerg Med, 2014. **47**(2): p. 131-9.
14. Rising, K.L., et al., *Patient returns to the emergency department: the time-to-return curve.* Acad Emerg Med, 2014. **21**(8): p. 864-71.
15. Keith, K.D., et al., *Emergency department revisits.* Ann Emerg Med, 1989. **18**(9): p. 964-8.
16. Lindsay, P., et al., *The Development of Indicators to Measure the Quality of Clinical Care in Emergency Departments Following a Modified-Delphi Approach.* Academic Emergency Medicine, 2002. **9**(11): p. 1131-1139.
17. Schenkel, S., *Promoting patient safety and preventing medical error in emergency departments.* Acad Emerg Med, 2000. **7**(11): p. 1204-22.
18. Pham, J.C., et al., *Seventy-two-hour returns may not be a good indicator of safety in the emergency department: a national study.* Acad Emerg Med, 2011. **18**(4): p. 390-7.
19. Abualenain, J., et al., *The prevalence of quality issues and adverse outcomes among 72-hour return admissions in the emergency department.* J Emerg Med, 2013. **45**(2): p. 281-8.
20. Lerman, B. and M.S. Kobernick, *Return visits to the emergency department.* J Emerg Med, 1987. **5**(5): p. 359-62.
21. Hu, K.W., et al., *Unscheduled return visits with and without admission post emergency department discharge.* J Emerg Med, 2012. **43**(6): p. 1110-8.
22. Calder, L., et al., *Adverse events in patients with return emergency department visits.* BMJ Qual Saf, 2015. **24**(2): p. 142-8.
23. Alessandrini, E.A., et al., *Return visits to a pediatric emergency department.* Pediatr Emerg Care, 2004. **20**(3): p. 166-71.
24. Chan, A.H., et al., *Characteristics of patients who made a return visit within 72 hours to the emergency department of a Singapore tertiary hospital.* Singapore Med J, 2016. **57**(6): p. 301-6.
25. Claret, P.G., et al., *Rates and predictive factors of return to the emergency department following an initial release by the emergency department for acute heart failure.* Cjem, 2017: p. 1-8.
26. Gaucher, N., B. Bailey, and J. Gravel, *Impact of physicians' characteristics on the admission risk among children visiting a pediatric emergency department.* Pediatr Emerg Care, 2012. **28**(2): p. 120-4.
27. Geirsson, O.P., et al., *Risk of repeat visits, hospitalisation and death after uncompleted and completed visits to the emergency department: a prospective observation study.* Emerg Med J, 2013. **30**(8): p. 662-8.
28. Gordon, J.A., et al., *Initial emergency department diagnosis and return visits: risk versus perception.* Ann Emerg Med, 1998. **32**(5): p. 569-73.
29. Martin-Gill, C. and R.C. Reiser, *Risk factors for 72-hour admission to the ED.* Am J Emerg Med, 2004. **22**(6): p. 448-53.
30. McCusker, J., et al., *Return to the emergency department among elders: patterns and predictors.* Acad Emerg Med, 2000. **7**(3): p. 249-59.
31. Pereira, L., et al., *Unscheduled-return-visits after an emergency department (ED) attendance and clinical link between both visits in patients aged 75 years and over: a prospective observational study.* PLoS One, 2015. **10**(4): p. e0123803.

32. van der Linden, M.C., et al., *Unscheduled return visits to a Dutch inner-city emergency department.* Int J Emerg Med, 2014. **7**: p. 23.
33. Vanbrabant, P. and D. Knockaert, *Short-term return visits of 'general internal medicine' patients to the emergency department: extent and risk factors.* Acta Clin Belg, 2009. **64**(5): p. 423-8.
34. Gabayan, G.Z., et al., *Factors associated with short-term bounce-back admissions after emergency department discharge.* Ann Emerg Med, 2013. **62**(2): p. 136-144.e1.
35. Sung, S.F., et al., *Predicting Factors and Risk Stratification for Return Visits to the Emergency Department Within 72 Hours in Pediatric Patients.* Pediatr Emerg Care, 2015. **31**(12): p. 819-24.
36. Suffoletto, B., et al., *Predicting older adults who return to the hospital or die within 30 days of emergency department care using the ISAR tool: subjective versus objective risk factors.* Emerg Med J, 2016. **33**(1): p. 4-9.
37. Lee, E.K., et al., *A clinical decision tool for predicting patient care characteristics: patients returning within 72 hours in the emergency department.* AMIA Annu Symp Proc, 2012. **2012**: p. 495-504.
38. LaMantia, M.A., et al., *Predicting hospital admission and returns to the emergency department for elderly patients.* Acad Emerg Med, 2010. **17**(3): p. 252-9.
39. Hao, S., et al., *Risk prediction of emergency department revisit 30 days post discharge: a prospective study.* PLoS One, 2014. **9**(11): p. e112944.
40. Graber, M.L., *Diagnostic errors in medicine: a case of neglect.* Jt Comm J Qual Patient Saf, 2005. **31**: p. 106-113.
41. Sabbatini, A.K., et al., *In-Hospital Outcomes and Costs Among Patients Hospitalized During a Return Visit to the Emergency Department.* Jama, 2016. **315**(7): p. 663-71.
42. Elixhauser, A., et al., *Clinical classifications for health policy research: hospital inpatient statistics, 1995.* Anonymous. Rockville, MD: Agency for Health Care Policy and Research, 1998: p. 98-0049.
43. Elixhauser, A. and C.A. Steiner, *Hospital inpatient statistics, 1996.* 1999: Diane Publishing Company.
44. Yang, C., et al. *Predicting 30-day all-cause readmissions from hospital inpatient discharge data.* in *e-Health Networking, Applications and Services (Healthcom), 2016 IEEE 18th International Conference on.* 2016. IEEE.
45. Lemke, K. *A predictive model to identify patients at risk of unplanned 30-day acute care hospital readmission.* in *Healthcare Informatics (ICHI), 2013 IEEE International Conference on.* 2013. IEEE.
46. Shulan, M., K. Gao, and C.D. Moore, *Predicting 30-day all-cause hospital readmissions.* Health care management science, 2013. **16**(2): p. 167-175.
47. Zhang, H., *Exploring conditions for the optimality of naive Bayes.* International Journal of Pattern Recognition and Artificial Intelligence, 2005. **19**(02): p. 183-198.
48. Agrawal, R. and R. Srikant. *Fast algorithms for mining association rules.* in *Proc. 20th int. conf. very large data bases, VLDB.* 1994.
49. Agrawal, D., et al., *Predicting Patients at Risk for 3-Day Postdischarge Readmissions, ED Visits, and Deaths.* Med Care, 2016. **54**(11): p. 1017-1023.
50. Shadmi, E., et al., *Predicting 30-day readmissions with preadmission electronic health record data.* Med Care, 2015. **53**(3): p. 283-9.
51. Donze, J., et al., *Potentially avoidable 30-day hospital readmissions in medical patients: derivation and validation of a prediction model.* JAMA Intern Med, 2013. **173**(8): p. 632-8.
52. van Walraven, C., et al., *Derivation and validation of an index to predict early death or unplanned readmission after discharge from hospital to the community.* Cmaj, 2010. **182**(6): p. 551-7.
53. Fleming, L.M., et al., *Derivation and validation of a 30-day heart failure readmission model.* Am J Cardiol, 2014. **114**(9): p. 1379-82.
54. Greenwald, J.L., et al., *A Novel Model for Predicting Rehospitalization Risk Incorporating Physical Function, Cognitive Status, and Psychosocial Support Using Natural Language Processing.* Med Care, 2017. **55**(3): p. 261-266.
55. Billings, J., et al., *Development of a predictive model to identify inpatients at risk of re-admission within 30 days of discharge (PARR-30).* BMJ Open, 2012. **2**(4).
56. Sun, Y., et al., *Predicting hospital admissions at emergency department triage using routine administrative data.* Acad Emerg Med, 2011. **18**(8): p. 844-50.
57. Kocher, K.E., J.B. Dimick, and B.K. Nallamothu, *Changes in the source of unscheduled hospitalizations in the United States.* Med Care, 2013. **51**(8): p. 689-98.
58. Bernstein, S.L., et al., *The effect of emergency department crowding on clinically oriented outcomes.* Acad Emerg Med, 2009. **16**(1): p. 1-10.
59. Ngai, K.M., et al., *The Association Between Limited English Proficiency and Unplanned Emergency Department Revisit Within 72 Hours.* Ann Emerg Med, 2016. **68**(2): p. 213-21.
60. Heitmann, M.G., et al., *Readmittance rates within seven days are preferable in quality measuring of emergency departments.* Dan Med J, 2016. **63**(9).